\pdfoutput=1

\documentclass[11pt]{article}

\usepackage{acl}

\usepackage{times}
\usepackage{latexsym}

\usepackage[T1]{fontenc}

\usepackage[utf8]{inputenc}

\usepackage{microtype}

\usepackage{multirow} 
\usepackage{booktabs} 
\usepackage{graphicx} 
\usepackage{bm} 
\usepackage{amsmath} 

\title{Bridge the Gap between  Language models and Tabular Understanding}

\author{Nuo Chen\textsuperscript{1}\thanks{\; Work done when interned at Microsoft STCA.},
  Linjun Shou\textsuperscript{2},
  Ming Gong\textsuperscript{2},
  Jian Pei\textsuperscript{3},
  Chenyu You\textsuperscript{4}\\
  \textbf{
  Jianhui Chang\textsuperscript{5},
  Daxin Jiang\textsuperscript{2},  Jia Li\textsuperscript{1}\thanks{\; Corresponding Author}} \\
  \textit{}
    \textsuperscript{\rm 1}Hong Kong University of Science and Technology (Guangzhou),\\ Hong Kong University of Science and Technology \\
	\textsuperscript{\rm 2}STCA, Microsoft, Beijing,
	\textsuperscript{\rm 3}Duke University, USA\\
 \textsuperscript{\rm 4}Yale University, USA,
 \textsuperscript{\rm 5}Peking University, China\\
\texttt{chennuo26@gmail.com}, \texttt{jialee@ust.hk} \\
}

\begin{document}
\maketitle
\begin{abstract}

Table pretrain-then-finetune paradigm has been proposed and employed at a rapid pace after the success of  pre-training in the natural language domain.
Despite the promising findings in tabular pre-trained language models (TPLMs), there is an input gap between pre-training and fine-tuning phases.
For instance, TPLMs jointly pre-trained with table and text input could be effective for tasks also with table-text joint input like table question answering, but it may fail
for tasks with only tables or text as input such as table retrieval. 
To this end, we propose UTP, an approach that dynamically supports three types of multi-modal inputs: table-text, table, and text. 
Specifically, UTP is pre-trained with two strategies: (1) We first utilize a universal mask language modeling objective on each kind of input, enforcing the model to 
adapt various inputs. (2) We then present Cross-Modal Contrastive Regularization (CMCR), which utilizes contrastive learning to encourage  the consistency  between  table-text cross-modality representations via unsupervised  instance-wise training signals during pre-training. By these means, the resulting model not only bridges the input gap between pre-training and fine-tuning but also advances in the alignment of table and text. Extensive results show UTP achieves superior results on uni-modal input tasks (e.g., table retrieval) and cross-modal input tasks (e.g., table question answering).

\end{abstract}

\section{Introduction}
Large-scale unsupervised pre-trained language models (PLMs) like BERT~\citep{devlin-etal-2019-bert} have been shown to be effective on various NLP tasks \cite{chen-etal-2022-bridging,chen2022good,you-etal-2022-end}. These models capture the syntactic and semantic information of natural language text during pre-training and can be adapted to various downstream tasks after fine-tuning. More recently, the pretrain-then-finetune paradigm has been extended to a new modality - Tabular.
\begin{table}[t]
\small
    \centering
    \begin{tabular}{ l c c}
    \toprule
    \textbf{Down-stream Task} & \textbf{Input}  & \textbf{Output}  \\
    \midrule
      Cell-Classification & Table & Label \\
     Table-Classification & Table & Label \\
    
 Table Retrieval &  Text & Table \\
    Table QA & Table + Text & Text \\
    Semantic Parsing & Table + Text & Text \\
    \bottomrule
    \end{tabular}
    \vspace{-2mm}
    \caption{The different data formats of downstream tasks.} 
    \label{table:data_format}
    \vspace{-3mm}
\end{table}
Various methods on tabular PLMs (TPLMs) have been proposed and successfully improved the performance of various table-related understanding tasks such as table/cell classification~\citep{deng2020turl,wang2021tuta,iida-etal-2021-tabbie} and table retrieval~\citep{herzig-etal-2020-tapas,yin-etal-2020-tabert,DBLP:conf/acl/HarariK22}. 

However, existing methods have two major limitations: 
First, they fix  input formats to match the specific downstream task during pre-training.
This may make the model sub-optimized in other downstream tasks, as shown in Table~\ref{table:data_format}.
For example, TAPAS~\citep{herzig-etal-2020-tapas} and TABERT~\citep{yin-etal-2020-tabert} require the pre-training with table-text pairs and formulate the inputs as concatenated table-text sequences, which are suitable for tasks like table QA, but perform unsatisfactorily for table-input only tasks like table classification and table filling~\citep{zhang2020web}.
In this paper, we call this phenomenon as the \textbf{input} gap in the pretrain-then-finetune  paradigm.
Second, their pre-training objectives are usually token-level, such as Masked Language Modeling (MLM) objective, cell-level cloze prediction and corrupt cell detection \cite{DBLP:conf/naacl/IidaTMI21}. Such objectives cannot capture the global alignment between table and text, which is critical for downstream tasks like table retrieval.

To address the above concerns, we propose \textbf{U}niversal \textbf{T}able-text \textbf{P}re-training (UTP), a novel table pre-training approach that can be adapted into various kinds of downstream tasks while capturing the global alignment between table and text. Specifically, we enable pre-training UTP to work on three different textual modalities: table, text, and table-text. We also apply mask language modeling (MLM) on each input to optimize the training. Furthermore, we design Cross-Modal Contrastive Regularization (CMCR)  to learn the alignment between table and text in a global manner. The core idea of CMCR is to push  representations of the same content in table and text to be similar, while pushing it away from other examples via contrastive learning \cite{you-etal-2021-self-supervised}.
Experimental results show that UTP outperforms  the state-of-the-art models for various table-related tasks such as table retrieval and table question answering. For instance, ours improves TAPAS from 28.95 to 38.45 in R@1 score on NQ-TABLES dataset.

\section{Related Work}
In recent years, more and more table pre-training methods \cite{yin-etal-2020-tabert,DBLP:conf/acl/YangGUHGP22,DBLP:conf/acl/Cheng0JWHCZ22,DBLP:conf/acl/YeL0S022} has been proposed for some particular table-related tasks. TABERT~\citep{yin-etal-2020-tabert} jointly learns representations for unstructured texts and structured tables by pre-training on millions of table-text pairs, and is successfully applied into tabular semantic parsing tasks. As a concurrent work, TAPAS~\citep{herzig-etal-2020-tapas} also focuses on the semantic parsing task but is trained from weak supervision and predicts denotations by selecting table cells and optionally applying a predicted aggregation operator to the selected cells. Besides, TURL~\citep{deng2020turl}, TUTA~\citep{wang2021tuta} and TABBIE~\citep{iida-etal-2021-tabbie} are proposed for table understanding tasks like table classification and table filling. Note that all the above models determine a particular input format that is suitable for the downstream tasks they focus on, such as pure table inputs or concatenated table-text inputs, while our UTP dynamically support table, text, and table-text sequences as inputs. For pre-training, all the above models adopt token-level objectives to algin the representations of table and text, such as masked language modeling, cell-level cloze prediction and corrupt cell detection, while we adopt a novel cross-modal contrastive objective to align table and text in a global manner.

\begin{figure*}[t] 
\vspace{-10pt}
\centerline{\includegraphics[width=0.9\linewidth]{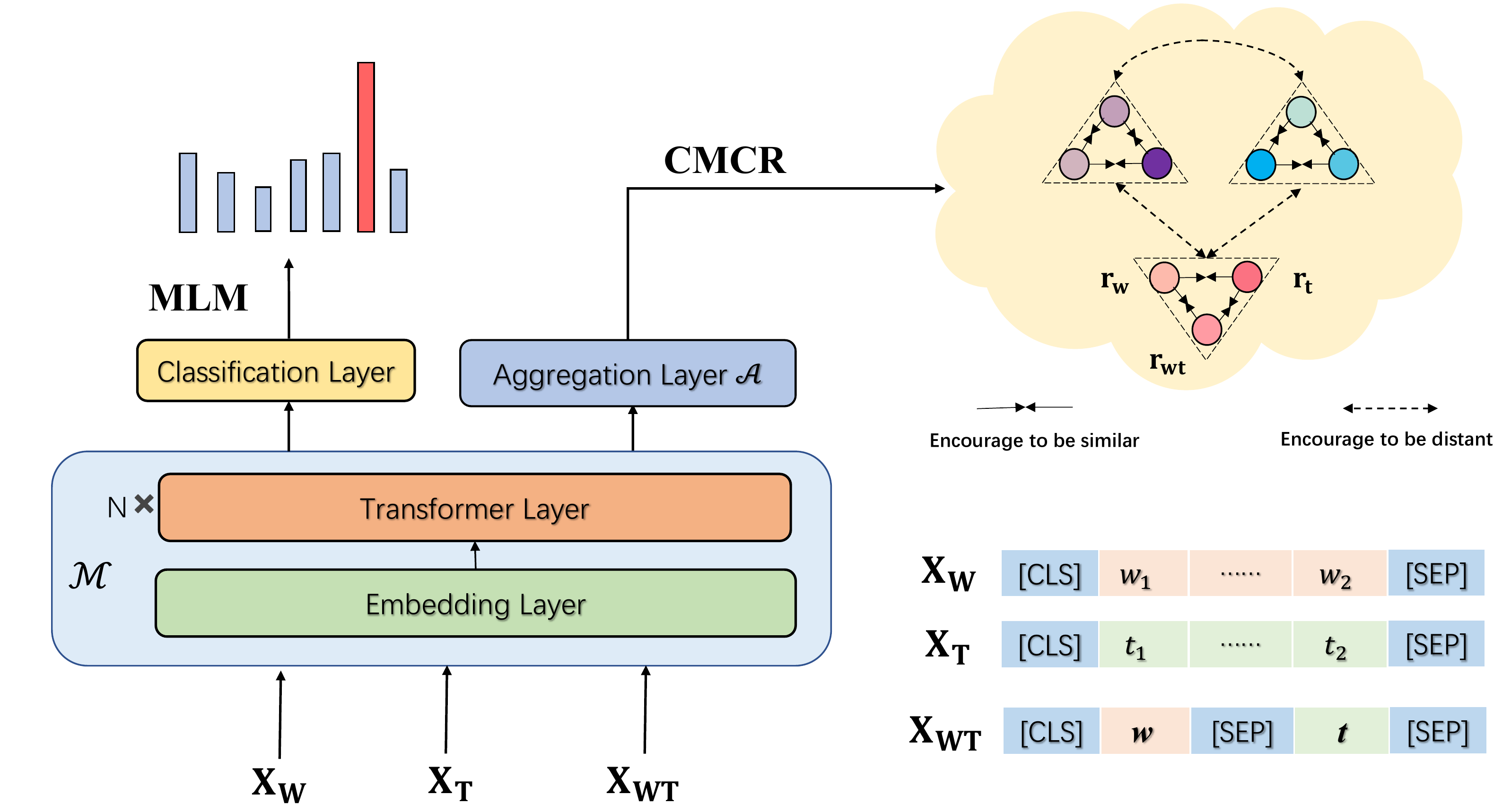}}
	\vspace{-2mm}
    \caption{Overview of our approach. UTP consists of three table-related inputs: $\mathbf{X_W}$, $\mathbf{X_T}$ and $\mathbf{X_{WT}}$. }
	\label{fig:utp_overview}
	\vspace{-3mm}
\end{figure*}


\section{Methodology}

    In this section, we present UTP and describe its detailed pre-training procedure.  As illustrated in Figure~\ref{fig:utp_overview}, UTP can operate on three different textual input modalities with a single unified model. Therefore, UTP can generalize well across various tasks.
    We first illustrate the input formats and universal mask language modeling in Section \ref{umlm}. Then we introduce the details of the proposed CMCR in Section \ref{cmcr}.
    
\subsection{Universal Mask Language Modeling}
\label{umlm}
 \paragraph{Model Architecture.} The designed model architecture of UTP: $\mathcal{M}$ is the same as TAPAS~\citep{herzig-etal-2020-tapas} which is based on BERT~\citep{devlin-etal-2019-bert} with seven additional positional embeddings for encoding tabular structure. 

\paragraph{Input Format.} Different from previous work~\citep{herzig-etal-2020-tapas,yin-etal-2020-tabert,iida-etal-2021-tabbie}, we do not restrict the input format to a particular form but allow for different input modalities, including a pure table input, a pure text input, and a joint table-text input. Formally, the input format of UTP can be summarized
as follows:
  \begin{align}
    &  \mathbf{X_W} = \{ \texttt{[CLS]} \bm{w}\texttt{[SEP]}\} \\
&\mathbf{X_T} = \{ \texttt{[CLS]} \bm{t} \texttt{[SEP]}\} \\
&\mathbf{X_{WT}} = \{ \texttt{[CLS]} \bm{w} \texttt{[SEP]} \bm{t} \texttt{[SEP]}\}, 
  \end{align}  
where  flattened table is $\bm{t}=\{t_1, \cdots, t_m\}$ and its corresponding text sequence refers to $\bm{w}=\{w_1, \cdots, w_n\}$. \texttt{[CLS]} and \texttt{[SEP]} denote the start  and separate tokens of the input sequence, separately.

    
During pre-training, we apply standard Masked Language Modeling (MLM) objective~\citep{devlin-etal-2019-bert} with 15\% of all word-piece tokens being masked in  both text and tabular data, as well as a combination of the two, expanding the range of forms it may accept. This allows the model to acquire not only a token-level comprehension of text and table, but also an alignment inside the two modals of presentation. Notice that  MLM losses of the three types of inputs are summed together as the final MLM objective: $\mathcal{L}_{mlm}$.
\subsection{Cross-Modal Contrastive Regularization}
\label{cmcr}
We design CMCR from the following perspectives: Misalignment between parallel sequences of table and text representations continues to prevent PLMs from progressing in downstream tabular tasks.
Thus, we present Cross-Modal Contrastive Regularization to align and unify the representations of texts and tables into the same semantic space. This method is beneficial for downstream tasks such as table-text retrieval, which need alignment between the two modalities, and developing more robust global sentence-level representations. As a result of the contrastive objective, the table and its surrounding text are located very close to one another in the representation space, as opposed to being located relatively far away in the case of an unpaired table and text. 

Specifically, we take $\mathbf{X_W}$, $\mathbf{X_T}$ and $\mathbf{X_{WT}}$ as inputs of $\mathcal{M}$ to get contextualized representations:
\begin{equation}
\small
\mathcal{H}_{w} = \mathcal{M}(\mathbf{X_W}), \quad \mathcal{H}_{t} = \mathcal{M}(\mathbf{X_T}), \quad
\mathcal{H}_{wt} = \mathcal{M}(\mathbf{X_{WT}}), 
\end{equation}
where $\mathcal{H}_{w} \in \mathbf{R}^{l\times d}$, $\mathcal{H}_{t} \in \mathbf{R}^{l\times d}$, $\mathcal{H}_{wt} \in \mathbf{R}^{l\times d}$, l and d are the  max sequence length of each input and hidden size, separately. Intuitively, we apply an extra aggregation layer (e.g., mean-pooling) $\mathcal{A}$  on $\mathcal{H}_{w}$, $\mathcal{H}_{t}$ and  $\mathcal{H}_{wt}$ to obtain final global semantics:
\begin{equation}
   \mathbf{r}_w = \mathcal{A}(\mathcal{H}_{w}), \quad \mathbf{r}_t = \mathcal{A}(\mathcal{H}_{t}), \quad
   \mathbf{r}_{wt} = \mathcal{A}(\mathcal{H}_{wt}),
\end{equation}
where $\mathbf{r}_w$, $\mathbf{r}_t$ and $\mathbf{r}_{wt}$ are in $\mathbf{R}^{d}$. Therefore, we can get positive threefold: $\{\mathbf{r}_{w}, \mathbf{r}_t, \hat{\mathbf{r}_{wt}} \}$. Each representation in this set is viewed as positive, which means that we anticipate each two of them to be as similar in the latent space as feasible. The others in the mini-batch are considered as negatives to optimize the $\mathcal{M}$ via a standard contrastive objective. For example, the cross-modal contrastive loss of ($\mathbf{r}_{t}$, $\mathbf{r}_{w}$) can be formulated as:
\begin{equation}
    \label{equation:taa_loss}
    \mathcal{L}(\mathbf{r}_{t}, \mathbf{r}_{w}) =  - \log \frac{\exp(\mathbf{r}_{t} \cdot \mathbf{r}_{w}) / \tau}{\sum_{j=1}^{N} \exp(\mathbf{r}_{t} \cdot \mathbf{r}_{w}^j) / \tau},
    \end{equation}
where N is the size of each mini-batch. $\cdot$ denotes inner product, and $\tau$ is a temperature parameter.
In general, the final loss of CMCR is as follows:
\begin{equation}
\mathcal{L}_{cmcr} = \mathcal{L}{(\mathbf{r}_{t}, \mathbf{r}_{w})} + \mathcal{L}{(\mathbf{r}_{t}, \mathbf{r}_{wt})} + \mathcal{L}{(\mathbf{r}_{wt}, \mathbf{r}_{w})}
\label{all_cons}
\end{equation}
Experimentally,  we optimize our model in a multi-task manner:
\begin{equation}
    \mathcal{L} = \mathcal{L}_{cmcr} + \mathcal{L}_{mlm}
\end{equation}

    

    





\section{Experiments}
In this section, we first describe the pre-training process including pre-training data and experimental setup. Then we introduce the downstream  datasets that are used to evaluate our model, as well as the fine-tuning details. Lastly, we report the experimental results on these datasets.

\subsection{Pre-training Settings}

\paragraph{Pre-training Data.} We collect tables and their surrounding text from English Wikipedia\footnote{\url{https://dumps.wikimedia.org/enwiki/}}. The data pre-processing procedure is same as TABERT~\citep{yin-etal-2020-tabert}. 
Each example in the pre-training data is a <\texttt{table}, \texttt{text}> pair where the text is the surrounding descriptions of the table.
The final pre-processed corpus contains 1.5 million parallel examples of tables and texts. The data is then randomly split into the train (90\%) and dev (10\%) sets.

\paragraph{Pre-training Details.} We use the TAPAS model in HuggingFace's Transformers library~\citep{wolf-etal-2020-transformers} as the base architecture of UTP and initialize UTP from TAPAS-Base. We pre-train UTP with 10 epochs on 8 Tesla V100 GPUs using 15 hours. Refer to Appendix  \ref{training_details} for more details of pre-training.

\subsection{Fine-tuning settings}

We fine-tune UTP on two categories of downstream tasks: (1) Uni-modal table retrieval task that requires the model to take text as input, aim at retrieving its corresponding table from a collection of tables;  (2) Cross-modal table question answering task, which takes unstructured text as query, and generates the answers from a source table. Current works in this task tend to concatenate text and table as input and then output the answer.

\paragraph{Datasets.} We finetune UTP on NQ-TABLES \cite{DBLP:conf/naacl/HerzigMKE21} and WIKISQL dataset.  NQ-TABLES \cite{DBLP:journals/corr/abs-1709-00103} is a well-famous table retrieval dataset that is derived from \cite{DBLP:journals/tacl/KwiatkowskiPRCP19}, which contains around 170k training corpora. WIKISQL is the largest human-annotated text-to-SQL tabular question answering dataset, which includes 80k QA pairs and 24k tables.
\paragraph{Fine-tuning Details.} Fine-tuning details are presented in Appendix \ref{training_details}, Table \ref{table:setup}.

\subsection{Results}

\paragraph{Table Retrieval} Table \ref{table:nqtable_results} shows the test results for table retrieval.
For each query, we record the recall at K (R@K) statistic as the proportion of the top K tables that match the reference. First, we can observe that all pre-trained dense-based models outperform the BM25 baseline by a large
margin. Second, UTP built on TAPAS also significantly surpasses its base model. For instance, ours improve TAPAS  by almost 10 points in R@1 and R@10. Third, we follow \cite{DBLP:conf/naacl/HerzigMKE21} to  extract hard negatives.  Moreover,  the addition of
mined negatives  yields an additional
improvement of each model. Similarly, UTP with hard negatives performs significantly better than other baselines, demonstrating the efficacy of our proposed methods once more (50.54 vs. 35.69 in R@1).

\begin{table}[t]
    \centering
    \resizebox{.47\textwidth}{!}{
    \begin{tabular}{l c c c}
    \toprule
    Model & R@1 & R@10 & R@50 \\
    \midrule
    BM25 \cite{DBLP:journals/ftir/RobertsonZ09} & 16.77 & 40.06 & 58.39 \\
    TAPAS & 28.95 & 69.71 & 86.38 \\
    DTR \cite{DBLP:conf/naacl/HerzigMKE21}& 35.64 & 76.14 & 91.43 \\
    \textbf{UTP}  & \textbf{38.45} & \textbf{79.03} & \textbf{92.21} \\
    \midrule
TAPAS + hn & 35.69 & 69.94 & 84.52 \\
   DTR + hnbm25 & 42.17 & 81.13 & 92.56 \\
    DTR + hn & 44.42 & 81.92 & 93.18 \\
    \textbf{UTP} + hn & \textbf{50.54} & \textbf{85.40} & \textbf{94.31} \\
    \bottomrule
    \end{tabular}
    }
    \caption{Table retrieval results on NQ-TABLES test set, averaged over 5 random runs. hn: hard negatives.
    } 
    \label{table:nqtable_results}
\end{table}

\begin{table}[t]
\footnotesize
    \centering
    
    \resizebox{.42\textwidth}{!}{
    \begin{tabular}{l c c}
    \toprule[0.8pt]
    Model &  Test \\
    \midrule

\cite{wang-etal-2019-learning} &   74.8 \\
 Hard-EM \cite{DBLP:conf/emnlp/MinCHZ19}   & 79.3 \\
    
    TAPAS &  83.6 \\
    \textbf{UTP} &  \textbf{84.5} \\
    \midrule
    
TAPAS (fully-supervised) & 86.4 \\
   \textbf{UTP} (fully-supervised) &  \textbf{88.1} \\
    \bottomrule[0.8pt]
    \end{tabular}
    }
    \caption{WIKISQL denotation accuracy.} 
    \label{table:wikisql_results}
    \vspace{-3mm}
\end{table}

\paragraph{Table QA.} We present experimental results for WIKISQL in Table \ref{table:wikisql_results}. Our UTP consistently achieves better performances than other baselines. Specifically, UTP obtains 88.1 (vs.86.4) when given the gold aggregation operators and cell as supervision (extracted from the reference SQL), which accounts for full supervision \cite{DBLP:conf/naacl/HerzigMKE21}. The results show that UTP can deal with different input modalities and achieve promising performances.
    
\subsection{Analysis}
\begin{table}[t]
    \centering
    \resizebox{.43\textwidth}{!}{
    \begin{tabular}{l c c c}
    \toprule
    Model & R@1 & R@10 & R@50 \\
    \midrule
  
    \textbf{UTP}  & \textbf{38.45} & \textbf{79.03} & \textbf{92.21} \\
    \midrule
\textit{w/o MLM} &  32.24 & 75.32 & 88.51 \\

 \textit{w/o CMCR} & 35.41 & 74.21 & 88.90 \\
\quad \textit{w/o} $\mathcal{L}{(\mathbf{r}_{t}, \mathbf{r}_{w})}$ & 36.08 & 75.06 & 89.66 \\
\quad \textit{w/o $\mathcal{L}{(\mathbf{r}_{t}, \mathbf{r}_{wt})}$} & 37.90 & 77.05 & 90.09 \\
\quad \textit{w/o $\mathcal{L}{(\mathbf{r}_{wt}, \mathbf{r}_{w})}$} & 37.64 & 77.62 & 90.64 \\
    \bottomrule
    \end{tabular}
    }
    \vspace{-2mm}
    \caption{Ablation study on NQ-TABLES test set, averaged over 5 random runs.
    } 
    \label{table:na_ablations}
\end{table}

\begin{figure}[t] 
	\centerline{\includegraphics[width=1\linewidth]{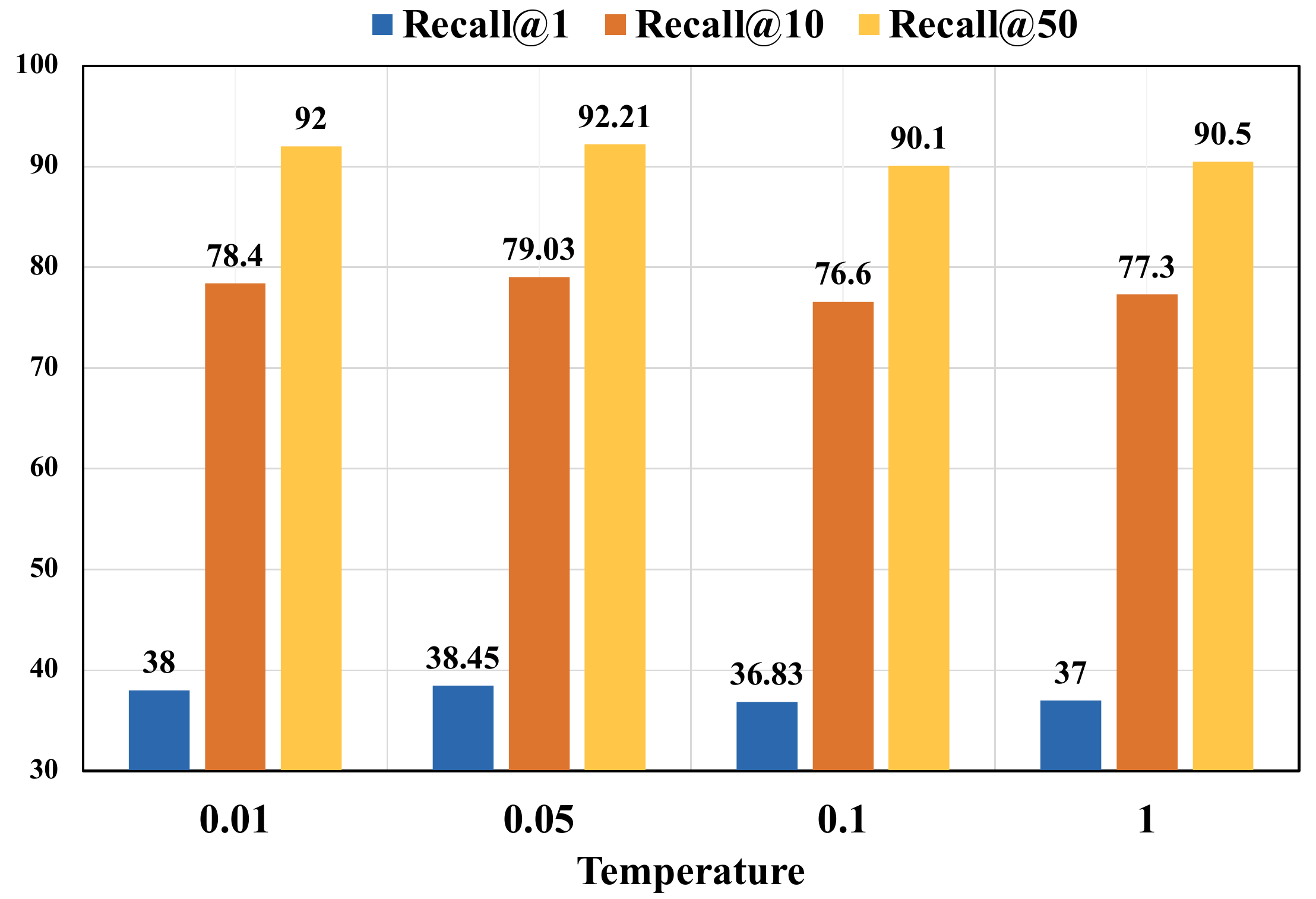}}
    \caption{Hyper-parameter analysis of $\tau$ in Eq.\ref{all_cons} on WIKISQL dataset.}
	\label{fig:utp_tem}
	\vspace{-3mm}
\end{figure}

\paragraph{Ablation Study.} In this subsection, we further conduct ablation studies to verify the effectiveness of each component in UTP. As seen in Table \ref{table:na_ablations}, both the  universal MLM and CMCR have a significant impact on the UTP performance enhancement, demonstrating the efficiency of the approach that we have proposed. For instance, removing MLM and CMCR could make the model performances decrease to 88.51 and 88.90 in R@50, separately. Besides, we also can observe that $\mathcal{L}{(\mathbf{r}_{t}, \mathbf{r}_{w})}$ loss plays the most important role in our proposed CMCR as it can directly align the modalities of pure text and table.

\paragraph{Hyper-parameter Analysis.} Intuitively, $\tau$ in Eq.\ref{all_cons} has the effect on our model's performances. 
Therefore, we conduct additional experiments to validate the influence of temperature $\tau$ on WIKISQL dataset.
We validate the model performances with $\tau \in \{0.01, 0.05, 0.1, 1\}$. As shown in Figure \ref{fig:utp_tem}, we show that our UTP obtains the optimal performance when $\tau=0.05$.


\section{Conclusion}
In this paper, we propose a novel framework (UTP) that can deal with various table-related tasks with a single unified model. We further propose CMCR to encourage table-text alignment at the sentence-level via contrastive learning.
With the intermediate table-text pre-training, UTP can perform well on both uni-modal (table) tasks and cross-modal (table-text) tasks. Results show that UTP obtains better  results compared with the current state-of-the-art models on several  table-related downstream tasks such as Table Retrieval and Table QA.
As an extension of our future work, it is interesting to extend UTP to multi-lingual scenarios by pre-training with cross-modal and cross-lingual tabular corpra.

\section*{Limitations}
The main target of this paper is to bridge the gap between pre-training and fine-tune phases for tabular pre-trained language  models (TPLMs). We present two straight  strategies to address this issue in the hope of boosting the performances of TPLMs. 
More generally, we expect our core idea can give some insights to other cross-modal tasks such as vision and text.
Admittedly, one limitation of this work is that the proposed methods need high-quality table-text pairs on a relatively large-scale.
These concerns warrant further research and consideration when utilizing this work to build effective tabular models.

\bibliography{anthology,custom}

\begin{thebibliography}{23}
\expandafter\ifx\csname natexlab\endcsname\relax\def\natexlab#1{#1}\fi

\bibitem[{Chen et~al.(2022{\natexlab{a}})Chen, Shou, Gong, and
  Pei}]{chen2022good}
Nuo Chen, Linjun Shou, Ming Gong, and Jian Pei. 2022{\natexlab{a}}.
\newblock From good to best: Two-stage training for cross-lingual machine
  reading comprehension.
\newblock In \emph{Proceedings of the AAAI Conference on Artificial
  Intelligence}, volume~36, pages 10501--10508.

\bibitem[{Chen et~al.(2022{\natexlab{b}})Chen, Shou, Gong, Pei, and
  Jiang}]{chen-etal-2022-bridging}
Nuo Chen, Linjun Shou, Ming Gong, Jian Pei, and Daxin Jiang.
  2022{\natexlab{b}}.
\newblock \href {https://doi.org/10.18653/v1/2022.naacl-main.139} {Bridging the
  gap between language models and cross-lingual sequence labeling}.
\newblock In \emph{Proceedings of the 2022 Conference of the North American
  Chapter of the Association for Computational Linguistics: Human Language
  Technologies}, pages 1909--1923, Seattle, United States. Association for
  Computational Linguistics.

\bibitem[{Cheng et~al.(2022)Cheng, Dong, Jia, Wu, Han, Cheng, and
  Zhang}]{DBLP:conf/acl/Cheng0JWHCZ22}
Zhoujun Cheng, Haoyu Dong, Ran Jia, Pengfei Wu, Shi Han, Fan Cheng, and Dongmei
  Zhang. 2022.
\newblock {FORTAP:} using formulas for numerical-reasoning-aware table
  pretraining.
\newblock In \emph{{ACL} {(1)}}, pages 1150--1166. Association for
  Computational Linguistics.

\bibitem[{Deng et~al.(2020)Deng, Sun, Lees, Wu, and Yu}]{deng2020turl}
Xiang Deng, Huan Sun, Alyssa Lees, You Wu, and Cong Yu. 2020.
\newblock Turl: Table understanding through representation learning.
\newblock \emph{arXiv preprint arXiv:2006.14806}.

\bibitem[{Devlin et~al.(2019)Devlin, Chang, Lee, and
  Toutanova}]{devlin-etal-2019-bert}
Jacob Devlin, Ming-Wei Chang, Kenton Lee, and Kristina Toutanova. 2019.
\newblock \href {https://doi.org/10.18653/v1/N19-1423} {{BERT}: Pre-training of
  deep bidirectional transformers for language understanding}.
\newblock In \emph{Proceedings of the 2019 Conference of the North {A}merican
  Chapter of the Association for Computational Linguistics: Human Language
  Technologies, Volume 1 (Long and Short Papers)}, pages 4171--4186,
  Minneapolis, Minnesota. Association for Computational Linguistics.

\bibitem[{Harari and Katz(2022)}]{DBLP:conf/acl/HarariK22}
Asaf Harari and Gilad Katz. 2022.
\newblock Few-shot tabular data enrichment using fine-tuned transformer
  architectures.
\newblock In \emph{{ACL} {(1)}}, pages 1577--1591. Association for
  Computational Linguistics.

\bibitem[{Herzig et~al.(2021)Herzig, M{\"{u}}ller, Krichene, and
  Eisenschlos}]{DBLP:conf/naacl/HerzigMKE21}
Jonathan Herzig, Thomas M{\"{u}}ller, Syrine Krichene, and Julian Eisenschlos.
  2021.
\newblock Open domain question answering over tables via dense retrieval.
\newblock In \emph{{NAACL-HLT}}, pages 512--519. Association for Computational
  Linguistics.

\bibitem[{Herzig et~al.(2020)Herzig, Nowak, M{\"u}ller, Piccinno, and
  Eisenschlos}]{herzig-etal-2020-tapas}
Jonathan Herzig, Pawel~Krzysztof Nowak, Thomas M{\"u}ller, Francesco Piccinno,
  and Julian Eisenschlos. 2020.
\newblock \href {https://doi.org/10.18653/v1/2020.acl-main.398} {{T}a{P}as:
  Weakly supervised table parsing via pre-training}.
\newblock In \emph{Proceedings of the 58th Annual Meeting of the Association
  for Computational Linguistics}, pages 4320--4333, Online. Association for
  Computational Linguistics.

\bibitem[{Iida et~al.(2021{\natexlab{a}})Iida, Thai, Manjunatha, and
  Iyyer}]{iida-etal-2021-tabbie}
Hiroshi Iida, Dung Thai, Varun Manjunatha, and Mohit Iyyer. 2021{\natexlab{a}}.
\newblock \href {https://doi.org/10.18653/v1/2021.naacl-main.270} {{TABBIE}:
  Pretrained representations of tabular data}.
\newblock In \emph{Proceedings of the 2021 Conference of the North American
  Chapter of the Association for Computational Linguistics: Human Language
  Technologies}, pages 3446--3456, Online. Association for Computational
  Linguistics.

\bibitem[{Iida et~al.(2021{\natexlab{b}})Iida, Thai, Manjunatha, and
  Iyyer}]{DBLP:conf/naacl/IidaTMI21}
Hiroshi Iida, Dung Thai, Varun Manjunatha, and Mohit Iyyer. 2021{\natexlab{b}}.
\newblock {TABBIE:} pretrained representations of tabular data.
\newblock In \emph{{NAACL-HLT}}, pages 3446--3456. Association for
  Computational Linguistics.

\bibitem[{Kwiatkowski et~al.(2019)Kwiatkowski, Palomaki, Redfield, Collins,
  Parikh, Alberti, Epstein, Polosukhin, Devlin, Lee, Toutanova, Jones, Kelcey,
  Chang, Dai, Uszkoreit, Le, and Petrov}]{DBLP:journals/tacl/KwiatkowskiPRCP19}
Tom Kwiatkowski, Jennimaria Palomaki, Olivia Redfield, Michael Collins,
  Ankur~P. Parikh, Chris Alberti, Danielle Epstein, Illia Polosukhin, Jacob
  Devlin, Kenton Lee, Kristina Toutanova, Llion Jones, Matthew Kelcey,
  Ming{-}Wei Chang, Andrew~M. Dai, Jakob Uszkoreit, Quoc Le, and Slav Petrov.
  2019.
\newblock Natural questions: a benchmark for question answering research.
\newblock \emph{Trans. Assoc. Comput. Linguistics}, 7:452--466.

\bibitem[{Min et~al.(2019)Min, Chen, Hajishirzi, and
  Zettlemoyer}]{DBLP:conf/emnlp/MinCHZ19}
Sewon Min, Danqi Chen, Hannaneh Hajishirzi, and Luke Zettlemoyer. 2019.
\newblock A discrete hard {EM} approach for weakly supervised question
  answering.
\newblock In \emph{{EMNLP/IJCNLP} {(1)}}, pages 2851--2864. Association for
  Computational Linguistics.

\bibitem[{Robertson and Zaragoza(2009)}]{DBLP:journals/ftir/RobertsonZ09}
Stephen~E. Robertson and Hugo Zaragoza. 2009.
\newblock The probabilistic relevance framework: {BM25} and beyond.
\newblock \emph{Found. Trends Inf. Retr.}, 3(4):333--389.

\bibitem[{Wang et~al.(2019)Wang, Titov, and Lapata}]{wang-etal-2019-learning}
Bailin Wang, Ivan Titov, and Mirella Lapata. 2019.
\newblock \href {https://doi.org/10.18653/v1/D19-1391} {Learning semantic
  parsers from denotations with latent structured alignments and abstract
  programs}.
\newblock In \emph{Proceedings of the 2019 Conference on Empirical Methods in
  Natural Language Processing and the 9th International Joint Conference on
  Natural Language Processing (EMNLP-IJCNLP)}, pages 3774--3785, Hong Kong,
  China. Association for Computational Linguistics.

\bibitem[{Wang et~al.(2021)Wang, Dong, Jia, Li, Fu, Han, and
  Zhang}]{wang2021tuta}
Zhiruo Wang, Haoyu Dong, Ran Jia, Jia Li, Zhiyi Fu, Shi Han, and Dongmei Zhang.
  2021.
\newblock Tuta: Tree-based transformers for generally structured table
  pre-training.
\newblock In \emph{Proceedings of the 27th ACM SIGKDD Conference on Knowledge
  Discovery \& Data Mining}, pages 1780--1790.

\bibitem[{Wolf et~al.(2020)Wolf, Debut, Sanh, Chaumond, Delangue, Moi, Cistac,
  Rault, Louf, Funtowicz, Davison, Shleifer, von Platen, Ma, Jernite, Plu, Xu,
  Le~Scao, Gugger, Drame, Lhoest, and Rush}]{wolf-etal-2020-transformers}
Thomas Wolf, Lysandre Debut, Victor Sanh, Julien Chaumond, Clement Delangue,
  Anthony Moi, Pierric Cistac, Tim Rault, Remi Louf, Morgan Funtowicz, Joe
  Davison, Sam Shleifer, Patrick von Platen, Clara Ma, Yacine Jernite, Julien
  Plu, Canwen Xu, Teven Le~Scao, Sylvain Gugger, Mariama Drame, Quentin Lhoest,
  and Alexander Rush. 2020.
\newblock \href {https://doi.org/10.18653/v1/2020.emnlp-demos.6} {Transformers:
  State-of-the-art natural language processing}.
\newblock In \emph{Proceedings of the 2020 Conference on Empirical Methods in
  Natural Language Processing: System Demonstrations}, pages 38--45, Online.
  Association for Computational Linguistics.

\bibitem[{Yang et~al.(2022)Yang, Gupta, Upadhyay, He, Goel, and
  Paul}]{DBLP:conf/acl/YangGUHGP22}
Jingfeng Yang, Aditya Gupta, Shyam Upadhyay, Luheng He, Rahul Goel, and Shachi
  Paul. 2022.
\newblock Tableformer: Robust transformer modeling for table-text encoding.
\newblock In \emph{{ACL} {(1)}}, pages 528--537. Association for Computational
  Linguistics.

\bibitem[{Ye et~al.(2022)Ye, Lin, Li, Sun, and Liu}]{DBLP:conf/acl/YeL0S022}
Deming Ye, Yankai Lin, Peng Li, Maosong Sun, and Zhiyuan Liu. 2022.
\newblock A simple but effective pluggable entity lookup table for pre-trained
  language models.
\newblock In \emph{{ACL} {(2)}}, pages 523--529. Association for Computational
  Linguistics.

\bibitem[{Yin et~al.(2020)Yin, Neubig, Yih, and Riedel}]{yin-etal-2020-tabert}
Pengcheng Yin, Graham Neubig, Wen-tau Yih, and Sebastian Riedel. 2020.
\newblock \href {https://doi.org/10.18653/v1/2020.acl-main.745} {{T}a{BERT}:
  Pretraining for joint understanding of textual and tabular data}.
\newblock In \emph{Proceedings of the 58th Annual Meeting of the Association
  for Computational Linguistics}, pages 8413--8426, Online. Association for
  Computational Linguistics.

\bibitem[{You et~al.(2022)You, Chen, Liu, Ge, Wu, and Zou}]{you-etal-2022-end}
Chenyu You, Nuo Chen, Fenglin Liu, Shen Ge, Xian Wu, and Yuexian Zou. 2022.
\newblock \href {https://doi.org/10.18653/v1/2022.findings-naacl.91}
  {End-to-end spoken conversational question answering: Task, dataset and
  model}.
\newblock In \emph{Findings of the Association for Computational Linguistics:
  NAACL 2022}, pages 1219--1232, Seattle, United States. Association for
  Computational Linguistics.

\bibitem[{You et~al.(2021)You, Chen, and Zou}]{you-etal-2021-self-supervised}
Chenyu You, Nuo Chen, and Yuexian Zou. 2021.
\newblock \href {https://doi.org/10.18653/v1/2021.findings-emnlp.3}
  {Self-supervised contrastive cross-modality representation learning for
  spoken question answering}.
\newblock In \emph{Findings of the Association for Computational Linguistics:
  EMNLP 2021}, pages 28--39, Punta Cana, Dominican Republic. Association for
  Computational Linguistics.

\bibitem[{Zhang and Balog(2020)}]{zhang2020web}
Shuo Zhang and Krisztian Balog. 2020.
\newblock Web table extraction, retrieval, and augmentation: A survey.
\newblock \emph{ACM Transactions on Intelligent Systems and Technology (TIST)},
  11(2):1--35.

\bibitem[{Zhong et~al.(2017)Zhong, Xiong, and
  Socher}]{DBLP:journals/corr/abs-1709-00103}
Victor Zhong, Caiming Xiong, and Richard Socher. 2017.
\newblock Seq2sql: Generating structured queries from natural language using
  reinforcement learning.
\newblock \emph{CoRR}, abs/1709.00103.

\end{thebibliography}
\bibliographystyle{acl_natbib}
\clearpage
\appendix

\section{Training details}
\label{training_details}
\begin{table}[ht]
    \centering
    {
    \begin{tabular}{l c c c}
    \toprule
   \textbf{Parameter} & Pre-training & NQ-TABLES & WIKISQL \\
    \midrule
  
    \textit{Batch size}  & 16 & 32 & 32 \\

\textit{Learning Rate} &  5e$^{-5}$ & 2e$^{-5}$ & 6.17e$^{-5}$ \\

\textit{Epoch} & 10 & 100 & 200 \\
\textit{Weight Decay} & 0.01 & 0.01 & 0.01 \\
\textit{Temperature} & 35.41 & - & - \\
\textit{fp16} & True & True & True \\
    \bottomrule
    \end{tabular}
    }
    \vspace{-2mm}
    \caption{Hyper-parameters setup during pre-training and fine-tuning.
    } 
    \label{table:setup}
    \vspace{-3mm}
\end{table}



\end{document}